\documentclass[conference]{IEEEtran}
\IEEEoverridecommandlockouts

\usepackage{cite}
\usepackage{amsmath,amssymb,amsfonts}
\usepackage{algorithmic}
\usepackage{graphicx}
\usepackage{textcomp}
\usepackage{tcolorbox}
\usepackage{xcolor}
\usepackage{verbatim}
\usepackage{booktabs}
\usepackage{url}
\def\BibTeX{{\rm B\kern-.05em{\sc i\kern-.025em b}\kern-.08em
    T\kern-.1667em\lower.7ex\hbox{E}\kern-.125emX}}
\begin{document}

\title{ESCUCHA: A Spanish Speech Benchmark for Heterogeneous Acoustic Conditions
}

\makeatletter
\newcommand{\linebreakand}{%
  \end{@IEEEauthorhalign}
  \hfill\mbox{}\par
  \mbox{}\hfill\begin{@IEEEauthorhalign}
}
\makeatother

    \author{
\IEEEauthorblockN{1\textsuperscript{st} Fernando López$^\dagger$}
\IEEEauthorblockA{\textit{Telefónica Innovación Digital}\\
\textit{Universidad Autónoma de Madrid}\\
Madrid, Spain \\
fernando.lopez@telefonica.com}
\and
\IEEEauthorblockN{2\textsuperscript{nd} Ana Ayala$^\dagger$}
\IEEEauthorblockA{\textit{Universidad Complutense de Madrid}\\
Madrid, Spain \\
aayala02@ucm.es}
\and
\IEEEauthorblockN{3\textsuperscript{rd} Guillermo Segovia}
\IEEEauthorblockA{\textit{Telefónica Innovación Digital}\\
Madrid, Spain \\
guillermo.segovia@telefonica.com}
\and
\IEEEauthorblockN{4\textsuperscript{th} Fernando Ibáñez}
\IEEEauthorblockA{\textit{Universidad Autónoma de Madrid}\\
Madrid, Spain \\
fernando.ibannez@estudiante.uam.es}
\and
\IEEEauthorblockN{5\textsuperscript{th} Ana Martínez}
\IEEEauthorblockA{\textit{Universidad Autónoma de Madrid}\\
Madrid, Spain \\
anamaria.martinezj@estudiante.uam.es}
\and
\IEEEauthorblockN{6\textsuperscript{th} Pablo Gómez}
\IEEEauthorblockA{\textit{Telefónica Innovación Digital}\\
Madrid, Spain \\
pablo.gomezguerrero@telefonica.com}
\linebreakand
\IEEEauthorblockN{7\textsuperscript{th} Jordi Luque}
\IEEEauthorblockA{\textit{Telefónica Innovación Digital}\\
Barcelona, Spain \\
jordi.luque@telefonica.com}
\and
\IEEEauthorblockN{~}
\IEEEauthorblockA{$^\dagger$These authors contributed equally to this work.}
}

\maketitle

\begin{abstract}
As large audio language models (LALMs) advance, robust evaluation frameworks have become essential. In this context, Spanish speech understanding under realistic acoustic conditions has received particularly little attention. We introduce ESCUCHA, the first Spanish speech understanding benchmark designed to evaluate LALMs across heterogeneous acoustic conditions and reasoning abilities. ESCUCHA comprises 1,000 human-curated questions paired with audio, totaling 162.9 hours sourced directly ``from the wild'' rather than drawn from existing datasets, with durations ranging from a few seconds to over 80 minutes. The benchmark emphasizes reasoning, spanning 9 perceptual and 10 reasoning categories, and it captures linguistic diversity through multiple Spanish accents and non-normative speech. ESCUCHA further includes multi-audio questions, spoken questions, and audio instructions, and it flags which questions support open-ended evaluation. Benchmarking several state-of-the-art multimodal and speech models reveals substantial performance gaps relative to trained humans.
\end{abstract}


\section{Introduction}
Understanding auditory information is essential to machine intelligence. Numerous large audio language models (LALMs) have emerged in response, including LTU \cite{gong2023listen}, SALMONN \cite{sun2024video}, GAMA \cite{ghosh2024gama}, Qwen3-Omni \cite{xu2025qwen3omnitechnicalreport}, Kimi-Audio \cite{ding2025kimi}, and Audio Flamingo 3 \cite{goel2025audioflamingo3advancing}.

Evaluating these models is critical for ranking their performance and exposing their limitations, and thereby advancing the field. Early benchmarks targeted foundational tasks such as speech recognition and speech translation \cite{huang2024dynamic, yang2024air, wang2024audiobench, huang2024dynamic2}. More recent efforts evaluate perception and reasoning jointly, including MMAU \cite{sakshi2024mmau}, MMAR \cite{ma2025mmar}, SAKURA \cite{yang2025sakura}, MMSU \cite{wang2025mmsu}, and MMAU-Pro \cite{kumar2026mmau}.

Despite this growth, it remains unclear whether strong performance on existing benchmarks transfers once models move beyond English and beyond normative speech. High scores under normative, English-centric conditions do not guarantee that a model will generalize to other languages and speaker characteristics, and current benchmarks offer little guidance on the expected degradation. This concern is acute for non-normative speech: neuromotor disorders such as amyotrophic lateral sclerosis (ALS) and post-stroke conditions frequently cause dysarthria, which impairs neuromuscular control of speech, reduces articulation clarity, alters prosody, and yields variable intelligibility, all of which challenge speech recognition systems \cite{rudzicz2012torgo, kim2008dysarthric}.

The gap is particularly concerning for practitioners who aim to deploy LALMs in non-English contexts. Multilingual benchmarks offer only limited relief: GlobeAudio spans six languages but excludes Spanish \cite{tan2026globeaudio}, while Fleurs-SLU \cite{schmidtfleurs} evaluates speech-LLMs on Spanish only as one language among many, and through a narrow set of reading-comprehension questions derived from text, whose answers are recoverable from a transcript. For Spanish specifically, initial efforts such as IberoBench \cite{baucells2025iberobench} target the Iberian languages, but IberoBench is a multi-task text benchmark for the natural language understanding of large language models (LLMs) rather than a speech benchmark. For non-normative speech, meanwhile, prior work centers on transcription, leaving speech understanding uncovered \cite{moure2026audio}. More broadly, existing benchmarks concentrate on English and normative speech, rely on read or curated audio rather than audio in the wild, restrict themselves to a limited range of question types, and, being derived from existing datasets, risk train-test contamination.

To close this gap, we introduce \textsc{ESCUCHA}, the first speech understanding benchmark for evaluating LALMs in Spanish, spanning both normative and non-normative speech sourced under real-world conditions. \textsc{ESCUCHA} follows the multiple-choice question answering (MCQA) protocol and additionally includes multi-audio comparative questions, audio instruction following (AIF), and spoken questions, which together characterize how LALMs perform on Spanish across diverse conditions. Our main contributions are as follows:
\begin{itemize}
    \item The first \textit{in-the-wild} Spanish benchmark (audio and questions) for evaluating large audio language models.\footnote{Download code and annotations: \url{https://github.com/ferugit/ESCUCHA}}
    \item The first LALM benchmark to evaluate speech understanding and reasoning over non-normative pathological speech, spanning amyotrophic lateral sclerosis (ALS) and stroke.
    \item A broad evaluation of current LALM behavior on Spanish, covering open-weight and closed models, text-only baselines, and human performance.
\end{itemize}

\section{The ESCUCHA Benchmark}
\label{sec:dataset} 

\subsection{Overview}
\label{ssec:overview}
\textsc{ESCUCHA} comprises 1{,}000 expert-curated questions grounded in 162.9 hours of in-the-wild Spanish audio sourced from publicly available recordings. Following the MMAU-Pro design philosophy~\cite{kumar2026mmau}, we annotate every question along two complementary axes: the \emph{perceptual} skills required to extract information from the signal, and the \emph{reasoning} skills required to transform that information into an answer, yielding a multi-label taxonomy of 9 perception and 10 reasoning categories. Questions carry 1.44 perception labels and 1.14 reasoning labels on average, reflecting the compositional nature of realistic audio understanding: a single item frequently demands, for example, both paralinguistic emotion recognition and causal reasoning about speaker intent. An example appears in Figure~\ref{fig:example-question}: the item combines non-normative speech from an ALS patient with temporal and quantitative reasoning, requiring the model to recognize two numerical values, place them on a shared timeline, and compare them.

\begin{figure}[h]
\centering
\begin{tcolorbox}[
  colback=gray!4!white,
  colframe=gray!55!black,
  title={\small\textsc{Example Question}},
  fonttitle=\bfseries\small,
  boxrule=0.55pt,
  arc=3pt,
  left=7pt, right=7pt, top=6pt, bottom=6pt
]
\small
\textbf{Audio context:} A 96-second testimony in which Irene, a 28-year-old woman with non-normative speech who was diagnosed with ALS (\textit{ELA}), recounts the date of her diagnosis and cites the life expectancy typically given to patients with the disease.

\medskip\noindent
\textbf{Question:}\\[2pt]
\textit{¿Desde que le diagnosticaron la ELA, ha vivido Irene más tiempo del que se
menciona en el vídeo como esperanza de vida para las personas con esta
enfermedad?}\\[1pt]
\textcolor{gray}{\footnotesize Since Irene was diagnosed with ALS, has she lived
longer than the life expectancy mentioned in the video for people with this disease?}

\medskip
\begin{tabular}{@{}c p{0.82\linewidth}@{}}
(A) & Ha vivido menos.
      \newline{\textcolor{gray}{\footnotesize\textit{She has lived less.}}}
      \\[5pt]
(B) & \textbf{Ha vivido más.}\enspace$\checkmark$
      \newline{\textcolor{gray}{\footnotesize\textit{She has lived longer.}}}
      \\[5pt]
(C) & No puede saberse.
      \newline{\textcolor{gray}{\footnotesize\textit{It cannot be determined.}}}
      \\[5pt]
(D) & Ha vivido lo mismo.
      \newline{\textcolor{gray}{\footnotesize\textit{She has lived the same amount of time.}}}
      \\
\end{tabular}

\smallskip
\hrule
\smallskip

\noindent\textit{Perception:}\enspace
  \textsf{Lexical \& Phrase-Level Recognition}\\
\noindent\textit{Reasoning:}\enspace
  \textsf{Quantitative Reasoning (Counting/Arithmetic Comparison)}
  $\cdot$
  \textsf{Temporal \& Ordering Reasoning}
\end{tcolorbox}

\caption{\textbf{Example of chained cross-category reasoning.} The question requires three steps: recognizing two numerical values (time since Irene's diagnosis and the life expectancy she cites), mapping them onto a shared timeline, and comparing them. The most frequent failure is selecting~(C)~\textit{``No puede saberse,''} which conflates the absence of an explicit comparison in the audio with genuine unknowability.}
\label{fig:example-question}
\end{figure}

ESCUCHA also features a \emph{comparative} subset: 254 questions (25.4\%) reference two separate recordings and require the model to relate them, comparing speakers, dialects, acoustic conditions, or content across clips. Such cross-clip comparison is common in professional listening tasks. Table~\ref{tab:summary} summarizes the main statistics of the benchmark.

\begin{table}[h]
  \centering
  \caption{ESCUCHA at a glance. AIF: audio instruction following;
  MCQA: multiple-choice question answering. Comparative, spoken, and non-normative are non-exclusive tags and do not partition the total.}
  \label{tab:summary}
  \begin{tabular}{lr}
    \toprule
    \textbf{Statistic} & \textbf{Value} \\
    \midrule
    Total questions                           & 1{,}000 \\
    \quad MCQA                                & 900 \\
    \quad AIF                                 & 100 \\
    \addlinespace
    \multicolumn{2}{l}{\emph{Non-exclusive questions}} \\
    \quad Comparative (two-audio)             & 254 \\
    \quad Spoken questions                    & 100 \\
    \quad Non-normative speech                & 158 \\
    \addlinespace
    Perception / reasoning categories         & 9 / 10 \\
    Mean labels per question (perc.\ / reas.) & 1.44 / 1.14 \\
    Duration per question (median / mean)     & 1.4 min / 9.8 min \\
    Duration range                            & 0.1 min -- 85.2 min \\
    \midrule
    Total audio (sum over questions)          & 162.9 h \\
    \bottomrule
  \end{tabular}
\end{table}

\subsection{Benchmark Creation Process}
\label{ssec:creation}
We construct ESCUCHA through the four-stage pipeline illustrated in Figure~\ref{fig:pipeline}.

\begin{figure*}[htp]
  \centering
  \includegraphics[width=0.7\textwidth]{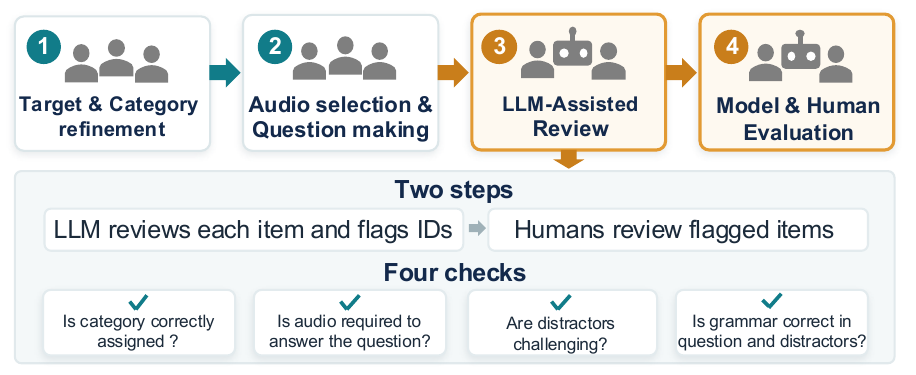}
  \caption{Pipeline for constructing the ESCUCHA benchmark. The process comprises four stages: (i)~target setting and refinement of the perception and reasoning categories; (ii)~audio selection and authoring of questions and distractors; (iii)~LLM-assisted review; and (iv)~model and human evaluation.}
  \label{fig:pipeline}
\end{figure*}

\begin{itemize}
    \item \textbf{Targets and category refinement.} We define the target of each question type, the categories, and the type of audio to collect, then revise the perception and reasoning taxonomy. Section~\ref{ssec:taxonomy} provides further detail.

    \item \textbf{Audio selection and question authoring.} We source audio from in-the-wild YouTube recordings, favoring spontaneous speech. Annotators identify candidate videos, filter the content, author the corresponding questions and distractors, ground each item in the audio, and assign its perception and reasoning labels. For non-normative speech, we retrieve videos using Spanish queries targeting dysarthric speech, such as condition-specific keywords, interviews, and documentaries, with a preference for interviews; we then filter candidates on the basis of self-reported diagnoses, video metadata, and perceptual evidence of non-normative speech. The resulting corpus spans acoustically diverse conditions, including background noise, music, and heterogeneous recording setups.

    \item \textbf{LLM-assisted review.} We review each question along four dimensions with an LLM (\texttt{Gemma-4-31B-IT}), which returns the identifiers of the items it flags for manual verification by annotators. The review checks whether the categories are correctly assigned, whether the audio is required to answer the question, whether the distractors are sufficiently challenging, and whether the question or distractors contain grammatical errors.

    \item \textbf{Model and human evaluation.} We benchmark a range of LALMs, text-only LLMs, and cascade systems, and we collect human judgments to establish reference performance. 
\end{itemize}
Across all four stages, the annotation team consists of three linguists and three technical experts, all native Spanish speakers. Each item carries metadata, including source URLs, measured durations, multi-label categories, distractors, and, where applicable, pathological speaker attributes and AIF verifier identifiers.

\subsection{Taxonomy}
\label{ssec:taxonomy}
We derive our taxonomy from MMAU-Pro~\cite{kumar2026mmau}, retaining its perception and reasoning axes while slightly redefining them. Table~\ref{tab:taxonomy} presents the category counts for our benchmark.

\begin{table}[t]
\centering
\small
\setlength{\tabcolsep}{4pt}
\caption{Question counts per perception and reasoning category. Counts sum to more than 1{,}000 because the taxonomy is multi-label.}
\begin{tabular}{@{}p{0.78\linewidth}r@{}}
\toprule
\textbf{Category} & \textbf{\#Q} \\
\midrule
\multicolumn{2}{@{}l}{\textit{Perception}} \\
\midrule
Lexical and Phrase-Level Recognition            & 506 \\
Speaker Identification                          & 324 \\
Paralinguistic/Emotion Recognition              & 181 \\
Speech Activity, Turn-Taking and Overlap Detection & 115 \\
Language Identification                         & 87  \\
Audio Quality, Artifacts \& Channel Characteristics & 84 \\
Prosody Detection                               & 74  \\
Speaker Demographics                            & 50  \\
Syntactic and Sentence-Structure Processing     & 18  \\
\midrule
\multicolumn{2}{@{}l}{\textit{Reasoning}} \\
\midrule
Speaker Intent, Pragmatics and Causal Reasoning & 221 \\
Semantic Abstraction and Summarization          & 177 \\
Logical/Consistency Reasoning                   & 159 \\
Ground Truth and World Knowledge Integration    & 133 \\
Quantitative Reasoning                          & 121 \\
Temporal and Ordering Reasoning                 & 105 \\
Social Role and Relationship Inference          & 94  \\
Comparative and Preference-Based Judgments      & 62  \\
Cross-frontier Entity Linking                   & 38  \\
Coherence and Discourse Structure               & 29  \\
\bottomrule
\end{tabular}
\label{tab:taxonomy}
\end{table}

Rather than enforce a uniform distribution, we prioritized questions that were challenging and well grounded in each recording over questions that would merely fill under-represented cells. This preserves difficulty while still covering every category. The co-occurrence matrix (Fig.~\ref{fig:cooc}) characterizes the distribution that emerged. It is densely populated rather than block-diagonal: lexical recognition co-occurs with all ten reasoning categories, peaking with semantic abstraction and summarization (158), while other perceptual skills pair more selectively, such as language identification with ground-truth and world-knowledge integration (53). This mix of one broad row and several localized pairings shows that many questions combine perception and reasoning rather than isolating a single skill. A few specific pairings remain unpopulated, which we leave to future work.

\begin{figure*}[t]
  \centering
  \includegraphics[width=0.75\textwidth]{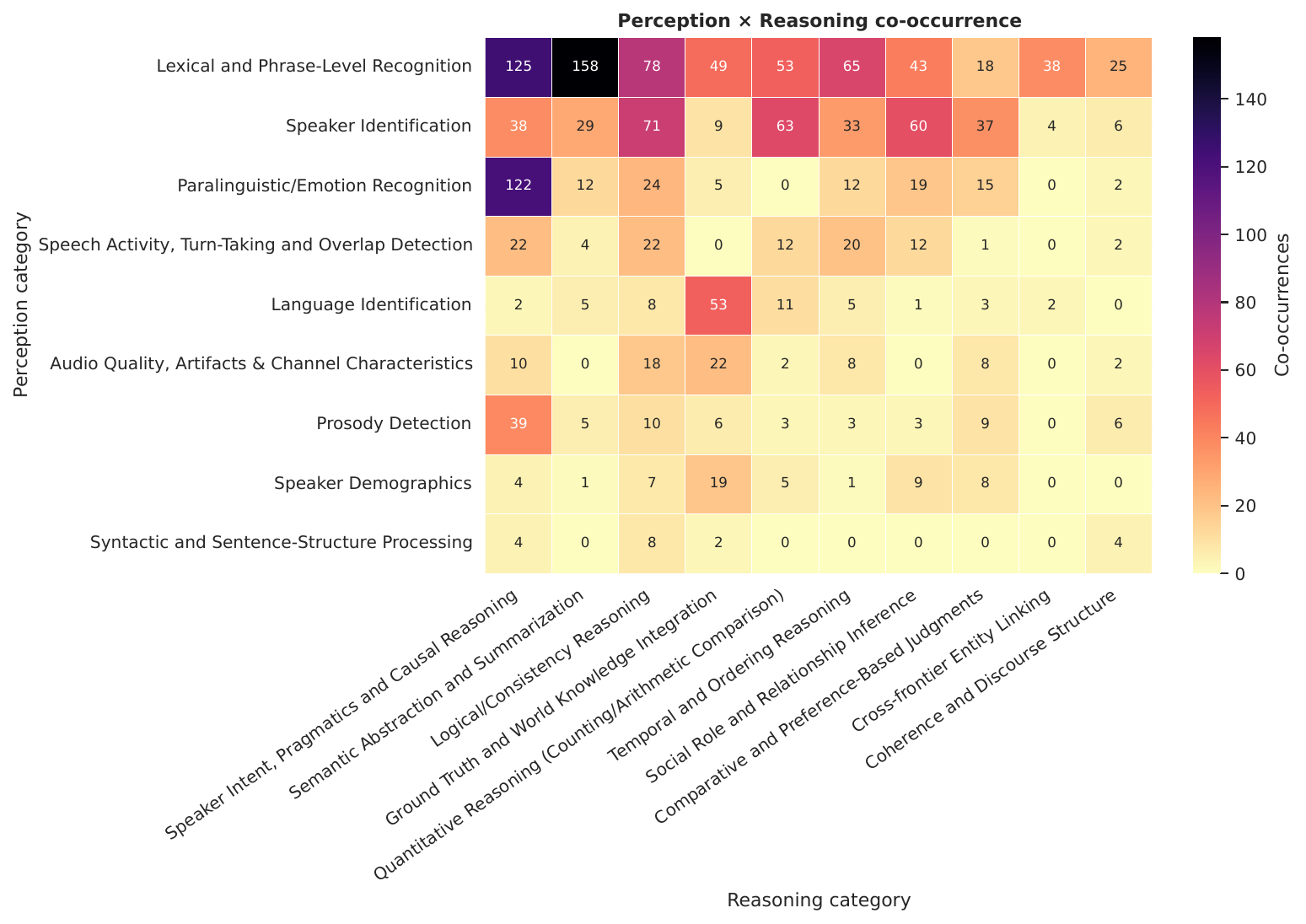}
  \vspace{-0.3cm}
  \caption{Perception~$\times$~reasoning co-occurrence across the 900 MCQA questions. AIF questions only with perception labels, were excluded. Each cell counts question label pairs; questions carrying multiple perception or reasoning labels contribute to several cells, so cell totals exceed the question count and row and column sums are not disjoint.}
  \label{fig:cooc}
\end{figure*}

\subsection{Other Tasks}
Beyond regular covered tasks, \textsc{ESCUCHA} includes two further subsets.

\textbf{Non-normative speech.}
A subset of 158 questions (15.8\%) targets pathological speech, with per-speaker metadata covering diagnosis, sex, and intelligibility. It includes speakers with ALS (114 questions) and post-stroke speech (44), annotated by sex (123 male, 35 female). We also provide intelligibility levels for 145 of the 158 items, ranging from Low (53) and Medium/Low (71) to Medium (9) and High/Medium (12); the remaining 13 ALS items lack an intelligibility annotation.

\textbf{Spoken questions.}
We sample 100 questions from the existing \textsc{ESCUCHA} pool, using inverse-frequency weighting over perception and reasoning categories to prioritise underrepresented classes. Only single-audio items were eligible, since the spoken question is stored as a second associated audio. The original question text is fed to the OmniVoice~\cite{zhu2026omnivoice} text-to-speech (TTS) system, which yields varied speakers and Spanish accents. At evaluation time, the written prompt is fixed (\textit{``Contesta a la pregunta en el audio''}, i.e., ``Answer the question in the audio''), so the model must recover the actual question entirely from the spoken audio. The choices are maintained in text.

\subsection{Audio Duration}
\label{ssec:duration}
The benchmark spans four orders of magnitude in duration, from 6.7~s to 85~min per question (median 86.2~s, mean 586.5~s; for comparative items the duration is summed over both clips). Each question is assigned to one of five length buckets, reported in Table~\ref{tab:duration}. The Long and Extended buckets jointly account for a quarter of the benchmark, allowing \textsc{ESCUCHA} to stress long-context audio understanding. Figure~\ref{fig:duration} shows the full duration distribution.

\begin{table}[t]
\centering
\small
\setlength{\tabcolsep}{6pt}
\caption{Distribution of questions across the five duration buckets.}
\begin{tabular}{@{}llrr@{}}
\toprule
\textbf{Bucket} & \textbf{Range} & \textbf{\#Q} & \textbf{\%} \\
\midrule
Brief    & ${\leq}30$~s        & 123 & 12.3 \\
Short    & 30~s--2~min         & 440 & 44.0 \\
Medium   & 2--10~min           & 187 & 18.7 \\
Long     & 10--30~min          & 127 & 12.7 \\
Extended & ${>}30$~min         & 123 & 12.3 \\
\midrule
\textbf{Total} &              & \textbf{1{,}000} & \textbf{100.0} \\
\bottomrule
\end{tabular}
\label{tab:duration}
\end{table}

\begin{figure}[t]
  \centering
  \includegraphics[width=\columnwidth]{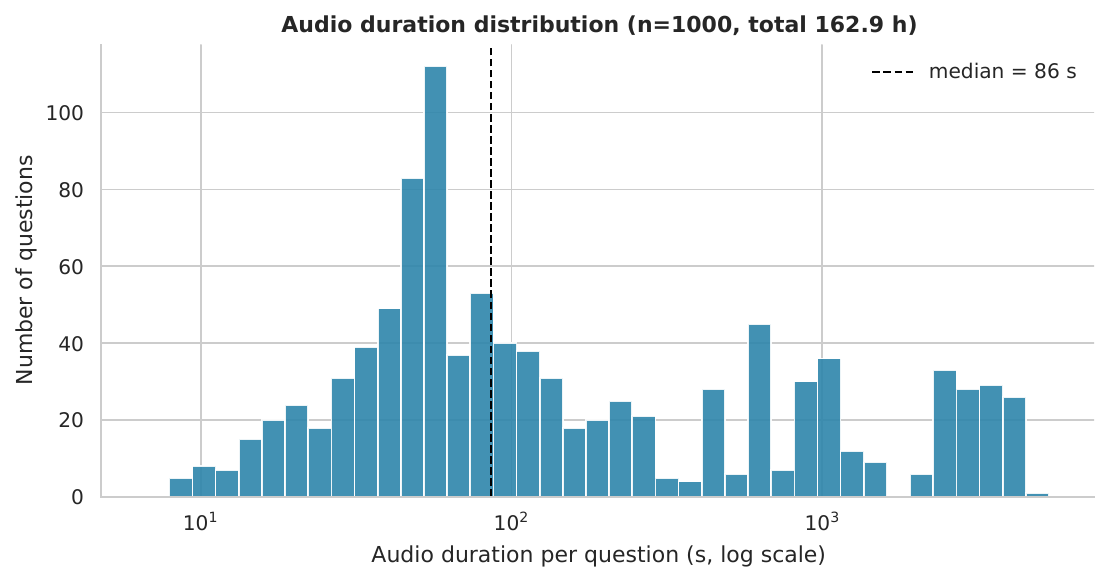}
  \caption{Distribution of audio duration per question (log scale). The dashed line marks the median (86.2~s).}
  \label{fig:duration}
\end{figure}

\subsection{Evaluation Formats}
\label{ssec:formats}
ESCUCHA contains two evaluation formats. Most of the benchmark (900 questions, 90\%) follows an MCQA protocol with 2--4 options (848 four-choice, 45 three-choice, 7 two-choice), enabling deterministic scoring. Independently of format, each question is annotated for whether the options are required: 712 (71.2\%) are answerable without them, allowing open-ended evaluation, and 288 (28.8\%) require them.

The remaining 100 questions (10\%) are \emph{audio instruction-following} (AIF) items: open-ended tasks in which the instruction is delivered in the audio rather than the text prompt, and the response must satisfy that spoken constraint, checked by a deterministic verifier. To produce the spoken instructions, we also use the OmniVoice~\cite{zhu2026omnivoice}. This design jointly probes audio comprehension and instruction following: the model must first recover the instruction from the signal and then obey it. We cover six constraint types:
\begin{itemize}
    \item \textbf{Keyword usage} (24): include a specified connector or phrase, e.g., \textit{sin embargo} (``however''), or at least one numeral.
    \item \textbf{Length} (22): meet an exact, minimum, maximum, or bounded word or sentence count, e.g., exactly three sentences.
    \item \textbf{Word exclusion} (18): avoid a specified word, expression, or punctuation mark, e.g., \textit{muy} (``very'').
    \item \textbf{Openings/closings} (16): begin with a fixed prefix or end with a fixed suffix, e.g., \textit{En resumen} (``In summary'').
    \item \textbf{Formatting} (12): adopt a structural form, e.g., a numbered list or sections delimited by \texttt{\#\#} headers.
    \item \textbf{Style} (8): e.g., write in lowercase only, pose the answer as questions, or repeat a keyword a number of times.
\end{itemize}
A constraint can occasionally be met without genuine compliance; word exclusion, for instance, may be satisfied by chance. The expected random-guessing accuracy is 25.61\% on the MCQA subset and 23.05\% on the full benchmark.

\section{Experimental Setup}
\label{sec:experimental_setup}
We evaluate \textsc{ESCUCHA} across four system families: end-to-end LALMs, a cascaded ASR-plus-LLM pipeline, text-only LLMs, and human and random baselines. Thus, we separate genuine audio understanding from what is recoverable through transcription alone and language bias. In our setup, multi-audio items are presented as their component clips concatenated with one second of silence between them, and all experiments run on an NVIDIA DGX Spark GPU.

\subsection{LALMs}
We evaluate several LALMs, covering open and proprietary systems of varying scale: \texttt{Audio-Flamingo-3} (8.3B)~\cite{goel2025audioflamingo3advancing}, \texttt{Qwen2.5-Omni} (7B), \texttt{Qwen3-Omni-30B-A3B}~\cite{xu2025qwen3omnitechnicalreport}, \texttt{Voxtral-Mini} (3B)~\cite{liu2025voxtral}, \texttt{Gemma-4-12B-IT}, and \texttt{Gemini-2.5-Flash}.

\subsection{Cascaded Systems}
To test whether audio understanding is required at all, or whether transcription suffices, we evaluate a cascade. This is a natural baseline given that lexical recognition is the most represented perception category. We transcribe each audio with \texttt{Whisper-Large-v3}~\cite{radford2023robust} and answer with \texttt{Qwen3-4B-Instruct-2507}~\cite{yang2025qwen3}.

\subsection{Text-Only LLMs}
Extending the cascade logic, we evaluate text-only LLMs that never see the raw audio. These receive the questions and choices with no audio. We include \texttt{Gemma-4-31B-IT}, \texttt{Gemma-4-12B-IT}, and \texttt{Qwen3-4B-Instruct-2507}~\cite{yang2025qwen3}.

\subsection{Evaluation Protocol}
\textbf{MCQA.}
Each item presents a question with $N$ labeled options; the model must respond as \texttt{<letter>.<justification>}. Scoring is exact-match: correct if the output matches the gold string case-insensitively or its first parsed letter indexes the gold option.

\textbf{AIF.}
Audio instruction-following items are open-ended: a deterministic \emph{verifier} function with a parameter dictionary is applied to the output, returning true or false. No model-based judge is used, so scoring is fully reproducible.

\section{Results}
\label{sec:results}
Table~\ref{tab:escucha_results} reports accuracy across all models and subsets. The trained human annotator reaches 90.10\% overall, far above the best model (74.40\%), establishing a wide human--model gap on \textsc{ESCUCHA}.

\begin{table*}[htp]
\centering
\caption{ESCUCHA benchmark results (accuracy \%). Best result per column is \textbf{bolded}. Pathological and Normative are MCQA subsets; Spoken Q contains items whose question is embedded in the audio.}
\label{tab:escucha_results}
\begin{tabular}{lcrrrrrr}
\toprule
\textbf{Model} & \textbf{Size} & \textbf{MCQA} & \textbf{AIF} & \textbf{Pathological} & \textbf{Normative} & \textbf{Spoken Q} & \textbf{Overall} \\
\midrule
\multicolumn{8}{c}{\textsc{Baselines}} \\
\midrule
\textit{Random} & -- & 25.61 & 0.00 & 25.32 & 25.67 & 25.75 & 23.05 \\
\textit{Always 1st} & -- & 26.22 & 0.00 & 24.05 & 26.68 & 23.00 & 23.60 \\
\textit{Always 2nd} & -- & 24.33 & 0.00 & 27.22 & 23.72 & 23.00 & 21.90 \\
\textit{Always 3rd} & -- & 26.22 & 0.00 & 25.95 & 26.28 & 30.00 & 23.60 \\
\textit{Always 4th} & -- & 23.22 & 0.00 & 22.78 & 23.32 & 24.00 & 20.90 \\
Human & -- & \textbf{91.33} & 79.00 & \textbf{93.04} & \textbf{90.97} & \textbf{93.00} & \textbf{90.10} \\
\midrule
\multicolumn{8}{c}{\textsc{Audio Models}} \\
\midrule
Audio-Flamingo-3-HF & 8.3B & 51.56 & 32.00 & 55.06 & 50.81 & 64.00 & 49.60 \\
Gemini-2.5-Flash & -- & 52.67 & 54.00 & 36.71 & 56.06 & 53.00 & 52.80 \\
Gemma-4-12B-IT & 12B & 48.11 & 60.00 & 36.08 & 50.67 & 49.00 & 49.30 \\
Qwen2.5-Omni-7B & 7B & 50.22 & 54.00 & 44.94 & 51.35 & 58.00 & 50.60 \\
Qwen3-Omni-30B-A3B & 30B (A3B) & 72.89 & \textbf{88.00} & 74.68 & 72.51 & 77.00 & 74.40 \\
Voxtral-Mini-3B-2507 & 3B & 50.22 & 40.00 & 51.90 & 49.87 & 55.00 & 49.20 \\
\midrule
\multicolumn{8}{c}{\textsc{Cascaded}} \\
\midrule
whisper-large-v3 + Qwen3-4B-Instruct-2507 & 4B & 59.00 & 69.00 & 63.92 & 57.95 & 59.00 & 60.00 \\
\midrule
\multicolumn{8}{c}{\textsc{Text-Only Models}} \\
\midrule
Gemma-4-31B-IT & 31B & 23.00 & 31.00 & 23.42 & 22.91 & 33.00 & 23.80 \\
Gemma-4-12B-IT (text) & 12B & 18.11 & 32.00 & 18.35 & 18.06 & 23.00 & 19.50 \\
Qwen3-4B-Instruct-2507 & 4B & 42.11 & 31.00 & 46.20 & 41.24 & 48.00 & 41.00 \\
\bottomrule
\end{tabular}
\end{table*}

Among audio models, \texttt{Qwen3-Omni-30B-A3B} stands out at 74.40\% overall and leads every column among models, including a striking 88.00\% on AIF that exceeds even the human score of 79.00\% on that subset. Every other audio model falls between 49 and 53\% overall, confirming that \textsc{ESCUCHA} poses a genuine challenge.

A consistent pattern is the drop from normative to pathological speech. \texttt{Gemini-2.5-Flash} and \texttt{Gemma-4-12B-IT} lose 19 and 15 percentage points, respectively, on pathological items, whereas \texttt{Audio-Flamingo-3}, \texttt{Qwen3-Omni-30B-A3B}, and \texttt{Voxtral-Mini} hold or slightly improve. This asymmetry indicates that pathological speech remains a weak point.

Comparison with MMAU-Pro~\cite{kumar2026mmau} adds context: \texttt{Audio-Flamingo-3} reaches 58.8\% on the MMAU-Pro speech subset but 49.60\% here, and \texttt{Qwen2.5-Omni-7B} drops similarly, from 57.4\% to 50.60\%. The two benchmarks are not directly comparable: they differ in question design, language, and the inclusion of pathological speech, but the gaps suggest \textsc{ESCUCHA} is at least as discriminative and harder for models with limited Spanish coverage.

The cascaded system (\texttt{Whisper-Large-v3} + \texttt{Qwen3-4B-Instruct-2507}) achieves 60.00\% overall, outperforming every single audio model except \texttt{Qwen3-Omni-30B-A3B}. Its advantage is most pronounced on AIF (69.00\%) and pathological items (63.92\%), supporting the view that a large fraction of \textsc{ESCUCHA} questions are grounded in lexical understanding.

All models receive the same system prompt, and none is instructed to guess under missing-modality conditions. The text-only \texttt{Qwen3-4B-Instruct-2507} reaches 41.00\% overall, well above the random baseline (23.05\%) and comparable to several audio systems, which we attribute to partial solvability from question structure and language priors. We therefore suggest treating this text-only model as a lower-bound baseline, alongside the positional controls, in future evaluations. The two text-only \texttt{Gemma-4} variants instead score near random (23.80\% and 19.50\%): lacking audio, they refuse a large fraction of items, 53\% for \texttt{Gemma-4-31B-IT} and 32\% for \texttt{Gemma-4-12B-IT}, with each refusal scored as incorrect and outputs such as requests to supply the audio for analysis.


\section{Limitations}
\label{sec:limitations}
Several factors qualify our results. The human upper bound is optimistic: the evaluator is a trained linguist, so the reported 90.10\% may differ from a regular person. On the AIF subset, some items can be answered correctly from the written instruction alone, without processing the audio, which partially inflates AIF scores and means the task does not exclusively measure audio instruction following. The pathological-speech subset is grounded in self-reported diagnoses targeting diverse speech understanding instead of clinical usage.

The evaluation is also uneven across models and axes. \texttt{Gemini-2.5-Flash} supports long-form audio architecturally, but a payload limit in our evaluation pipeline forced us to exclude items above a size threshold, biasing its results toward shorter items and preventing a direct comparison with models run on the full set. More broadly, \textsc{ESCUCHA} covers speech exclusively, so results do not generalize to music, environmental sound, or audio effects, and the benchmark is constructed entirely in Spanish, meaning scores for models with limited Spanish data conflate audio understanding with language coverage. Cross-lingual comparisons against English-centric benchmarks such as MMAU-Pro should be made with care. Finally, the taxonomy is lexically unbalanced, with lexical and phrase-level recognition dominating the perception axis, so per-category scores in sparsely populated cells rest on few items.

\section{Conclusions}
We introduced \textsc{ESCUCHA}, the first in-the-wild Spanish speech benchmark for large audio-language models, and the first to evaluate reasoning over non-normative pathological speech. It pairs 1{,}000 curated questions with 162.9 hours of audio, annotated along a multi-label taxonomy of nine perception and ten reasoning categories, and combines MCQA with comparative multi-audio items, audio instruction following, and spoken questions.

Our evaluation reveals a wide human--model gap: the best system reaches 74.40\% overall against 90.10\% for a trained annotator, with most audio models near 50\%. Two findings stand out. Several models degrade markedly on pathological speech, a persistent weak point for LALMs trained on clean, read audio. And a cascaded ASR-plus-LLM pipeline outperforms every single audio model but one, indicating that much of the benchmark is grounded in lexical understanding recoverable from a transcript, while the residual gap to \texttt{Qwen3-Omni-30B-A3B} marks where genuine acoustic reasoning still matters.

Future work includes broadening the pathological subset with further neurological conditions, and populating the currently sparse perception-reasoning pairings.

\section*{AI-Generated Content Disclosure}
We used a generative AI to assist in paraphrasing and improving clarity and grammar in parts of the manuscript; all generated content was reviewed and validated by the authors.


\begin{thebibliography}{10}
\providecommand{\url}[1]{#1}
\csname url@samestyle\endcsname
\providecommand{\newblock}{\relax}
\providecommand{\bibinfo}[2]{#2}
\providecommand{\BIBentrySTDinterwordspacing}{\spaceskip=0pt\relax}
\providecommand{\BIBentryALTinterwordstretchfactor}{4}
\providecommand{\BIBentryALTinterwordspacing}{\spaceskip=\fontdimen2\font plus
\BIBentryALTinterwordstretchfactor\fontdimen3\font minus \fontdimen4\font\relax}
\providecommand{\BIBforeignlanguage}[2]{{%
\expandafter\ifx\csname l@#1\endcsname\relax
\typeout{** WARNING: IEEEtran.bst: No hyphenation pattern has been}%
\typeout{** loaded for the language `#1'. Using the pattern for}%
\typeout{** the default language instead.}%
\else
\language=\csname l@#1\endcsname
\fi
#2}}
\providecommand{\BIBdecl}{\relax}
\BIBdecl

\bibitem{gong2023listen}
Y.~Gong, H.~Luo, A.~H. Liu, L.~Karlinsky, and J.~Glass, ``Listen, think, and understand,'' in \emph{Proc. ICLR}, 2024.

\bibitem{sun2024video}
G.~Sun, W.~Yu, C.~Tang, X.~Chen, T.~Tan, W.~Li, L.~Lu, Z.~Ma, Y.~Wang, and C.~Zhang, ``video-salmonn: Speech-enhanced audio-visual large language models,'' in \emph{Proc. ICML}, 2024.

\bibitem{ghosh2024gama}
\BIBentryALTinterwordspacing
S.~Ghosh, S.~Kumar, A.~Seth, C.~K.~R. Evuru, U.~Tyagi, S.~Sakshi, O.~Nieto, R.~Duraiswami, and D.~Manocha, ``{GAMA: A large audio-language model with advanced audio understanding and complex reasoning abilities},'' in \emph{Proc. EMNLP}.\hskip 1em plus 0.5em minus 0.4em\relax Miami, Florida, USA: ACL, 2024, pp. 6288--6313. [Online]. Available: \url{https://aclanthology.org/2024.emnlp-main.361/}
\BIBentrySTDinterwordspacing

\bibitem{xu2025qwen3omnitechnicalreport}
\BIBentryALTinterwordspacing
J.~Xu, Z.~Guo, H.~Hu, Y.~Chu, X.~Wang, J.~He, Y.~Wang, X.~Shi, T.~He, X.~Zhu, Y.~Lv, Y.~Wang, D.~Guo, H.~Wang, L.~Ma, P.~Zhang, X.~Zhang, H.~Hao, Z.~Guo, B.~Yang, B.~Zhang, Z.~Ma, X.~Wei, S.~Bai, K.~Chen, X.~Liu, P.~Wang, M.~Yang, D.~Liu, X.~Ren, B.~Zheng, R.~Men, F.~Zhou, B.~Yu, J.~Yang, L.~Yu, J.~Zhou, and J.~Lin, ``Qwen3-omni technical report,'' 2025. [Online]. Available: \url{https://arxiv.org/abs/2509.17765}
\BIBentrySTDinterwordspacing

\bibitem{ding2025kimi}
D.~Ding, Z.~Ju, Y.~Leng, S.~Liu, T.~Liu, Z.~Shang, K.~Shen, W.~Song, X.~Tan, H.~Tang \emph{et~al.}, ``Kimi-audio technical report,'' \emph{arXiv preprint arXiv:2504.18425}, 2025.

\bibitem{goel2025audioflamingo3advancing}
\BIBentryALTinterwordspacing
S.~Ghosh, A.~Goel, J.~Kim, S.~Kumar, Z.~Kong, S.~gil Lee, C.-H.~H. Yang, R.~Duraiswami, D.~Manocha, R.~Valle, and B.~Catanzaro, ``Audio flamingo 3: Advancing audio intelligence with fully open large audio language models,'' in \emph{Proc. Adv. Neural Inf. Process. Syst. (NeurIPS)}, 2025. [Online]. Available: \url{https://openreview.net/forum?id=FjByDpDVIO}
\BIBentrySTDinterwordspacing

\bibitem{huang2024dynamic}
C.-y. Huang, K.-H. Lu, S.-H. Wang, C.-Y. Hsiao, C.-Y. Kuan, H.~Wu, S.~Arora, K.-W. Chang, J.~Shi, Y.~Peng \emph{et~al.}, ``Dynamic-superb: Towards a dynamic, collaborative, and comprehensive instruction-tuning benchmark for speech,'' in \emph{Proc. ICASSP}.\hskip 1em plus 0.5em minus 0.4em\relax IEEE, 2024, pp. 12\,136--12\,140.

\bibitem{yang2024air}
\BIBentryALTinterwordspacing
Q.~Yang, J.~Xu, W.~Liu, Y.~Chu, Z.~Jiang, X.~Zhou, Y.~Leng, Y.~Lv, Z.~Zhao, C.~Zhou \emph{et~al.}, ``Air-bench: Benchmarking large audio-language models via generative comprehension,'' in \emph{Proc. ACL}.\hskip 1em plus 0.5em minus 0.4em\relax Bangkok, Thailand: ACL, Aug. 2024, pp. 1979--1998. [Online]. Available: \url{https://aclanthology.org/2024.acl-long.109/}
\BIBentrySTDinterwordspacing

\bibitem{wang2024audiobench}
B.~Wang, X.~Zou, G.~Lin, S.~Sun, Z.~Liu, W.~Zhang, Z.~Liu, A.~Aw, and N.~F. Chen, ``Audiobench: A universal benchmark for audio large language models,'' in \emph{Proc. NAACL:HLT}, Albuquerque, New Mexico, 2025, pp. 4297--4316.

\bibitem{huang2024dynamic2}
C.-y. Huang, W.-C. Chen, S.-w. Yang, A.~T. Liu, C.-A. Li, Y.-X. Lin, W.-C. Tseng, A.~Diwan, Y.-J. Shih, J.~Shi \emph{et~al.}, ``Dynamic-superb phase-2: A collaboratively expanding benchmark for measuring the capabilities of spoken language models with 180 tasks,'' in \emph{Proc. ICLR}, 2025.

\bibitem{sakshi2024mmau}
S.~Sakshi, U.~Tyagi, S.~Kumar, A.~Seth, R.~Selvakumar, O.~Nieto, R.~Duraiswami, S.~Ghosh, and D.~Manocha, ``{MMAU: A massive multi-task audio understanding and reasoning benchmark},'' in \emph{Proc. ICLR}, 2024.

\bibitem{ma2025mmar}
Z.~Ma, Y.~Ma, Y.~Zhu, C.~Yang, Y.-W. Chao, R.~Xu, W.~Chen, Y.~Chen, Z.~Chen, J.~Cong \emph{et~al.}, ``Mmar: A challenging benchmark for deep reasoning in speech, audio, music, and their mix,'' in \emph{Proc. Adv. Neural Inf. Process. Syst. (NeurIPS) Datasets and Benchmarks Track}, 2025.

\bibitem{yang2025sakura}
C.-K. Yang, N.~Ho, Y.-T. Piao, and H.~yi~Lee, ``{SAKURA: On the Multi-hop Reasoning of Large Audio-Language Models Based on Speech and Audio Information},'' in \emph{{Interspeech}}, 2025, pp. 1788--1792.

\bibitem{wang2025mmsu}
D.~Wang, J.~Wu, J.~Li, D.~Yang, X.~Chen, T.~Zhang, and H.~Meng, ``{MMSU: A Massive Multi-task Spoken Language Understanding and Reasoning Benchmark},'' \emph{arXiv preprint arXiv:2506.04779}, 2025.

\bibitem{kumar2026mmau}
S.~Kumar, {\v{S}}.~Sedl{\'a}{\v{c}}ek, V.~Lokegaonkar, F.~L{\'o}pez, W.~Yu, N.~Anand, H.~Ryu, L.~Chen, M.~Pli{\v{c}}ka, M.~Hlav{\'a}{\v{c}}ek \emph{et~al.}, ``Mmau-pro: A challenging and comprehensive benchmark for holistic evaluation of audio general intelligence,'' in \emph{Proc. AAAI Conf. Artif. Intell. (AAAI)}, vol.~40, no.~27, 2026, pp. 22\,688--22\,697.

\bibitem{rudzicz2012torgo}
F.~Rudzicz, A.~K. Namasivayam, and T.~Wolff, ``The torgo database of acoustic and articulatory speech from speakers with dysarthria,'' \emph{Lang. Resources Eval.}, vol.~46, no.~4, pp. 523--541, 2012.

\bibitem{kim2008dysarthric}
H.~Kim, M.~Hasegawa-Johnson, A.~Perlman, J.~R. Gunderson, T.~S. Huang, K.~L. Watkin, S.~Frame \emph{et~al.}, ``Dysarthric speech database for universal access research.'' in \emph{Interspeech}, vol. 2008, 2008, pp. 1741--1744.

\bibitem{tan2026globeaudio}
R.~Tan and W.~Zhang, ``Globeaudio: A multilingual multicultural benchmark for naturalistic evaluation of large audio-language models,'' \emph{arXiv preprint arXiv:2606.08194}, 2026.

\bibitem{schmidtfleurs}
F.~D. Schmidt, I.~Vuli{\'c}, G.~Glava{\v{s}}, and D.~I. Adelani, ``Fleurs-slu: A massively multilingual benchmark for spoken language understanding,'' in \emph{Proc. Conf. Lang. Model. (COLM)}, 2025.

\bibitem{baucells2025iberobench}
I.~Baucells, J.~Aula-Blasco, I.~de~Dios-Flores, S.~P. Su{\'a}rez, N.~Perez, A.~Salles, S.~S. Docio, J.~Falc{\~a}o, J.~J. Saiz, R.~Sep{\'u}lveda-Torres \emph{et~al.}, ``Iberobench: A benchmark for llm evaluation in iberian languages,'' in \emph{Proc. Int. Conf. Comput. Linguistics (COLING)}, 2025, pp. 10\,491--10\,519.

\bibitem{moure2026audio}
P.~Moure, N.~Pokel, B.~Bounajma, Y.~Gao, R.~Boehringer, L.~Cheng, and S.-C. Liu, ``When audio-language models fail to leverage multimodal context for dysarthric speech recognition,'' \emph{arXiv preprint arXiv:2605.02782}, 2026.

\bibitem{zhu2026omnivoice}
H.~Zhu, L.~Ye, W.~Kang, Z.~Yao, L.~Guo, F.~Kuang, Z.~Han, W.~Zhuang, L.~Lin, and D.~Povey, ``Omnivoice: Towards omnilingual zero-shot text-to-speech with diffusion language models,'' \emph{arXiv preprint arXiv:2604.00688}, 2026.

\bibitem{liu2025voxtral}
A.~H. Liu, A.~Ehrenberg, A.~Lo, C.~Denoix, C.~Barreau, G.~Lample, J.-M. Delignon, K.~R. Chandu, P.~von Platen, P.~R. Muddireddy \emph{et~al.}, ``Voxtral,'' \emph{arXiv preprint arXiv:2507.13264}, 2025.

\bibitem{radford2023robust}
A.~Radford, J.~W. Kim, T.~Xu, G.~Brockman, C.~McLeavey, and I.~Sutskever, ``Robust speech recognition via large-scale weak supervision,'' in \emph{Proc. ICML}.\hskip 1em plus 0.5em minus 0.4em\relax PMLR, 2023, pp. 28\,492--28\,518.

\bibitem{yang2025qwen3}
A.~Yang, A.~Li, B.~Yang, B.~Zhang, B.~Hui, B.~Zheng, B.~Yu, C.~Gao, C.~Huang, C.~Lv \emph{et~al.}, ``Qwen3 technical report,'' \emph{arXiv preprint arXiv:2505.09388}, 2025.

\end{thebibliography}

\begin{thebibliography}{00}
\bibitem{b1} G. Eason, B. Noble, and I. N. Sneddon, ``On certain integrals of Lipschitz-Hankel type involving products of Bessel functions,'' Phil. Trans. Roy. Soc. London, vol. A247, pp. 529--551, April 1955.
\bibitem{b2} J. Clerk Maxwell, A Treatise on Electricity and Magnetism, 3rd ed., vol. 2. Oxford: Clarendon, 1892, pp.68--73.
\bibitem{b3} I. S. Jacobs and C. P. Bean, ``Fine particles, thin films and exchange anisotropy,'' in Magnetism, vol. III, G. T. Rado and H. Suhl, Eds. New York: Academic, 1963, pp. 271--350.
\bibitem{b4} K. Elissa, ``Title of paper if known,'' unpublished.
\bibitem{b5} R. Nicole, ``Title of paper with only first word capitalized,'' J. Name Stand. Abbrev., in press.
\bibitem{b6} Y. Yorozu, M. Hirano, K. Oka, and Y. Tagawa, ``Electron spectroscopy studies on magneto-optical media and plastic substrate interface,'' IEEE Transl. J. Magn. Japan, vol. 2, pp. 740--741, August 1987 [Digests 9th Annual Conf. Magnetics Japan, p. 301, 1982].
\bibitem{b7} M. Young, The Technical Writer's Handbook. Mill Valley, CA: University Science, 1989.
\bibitem{b8} D. P. Kingma and M. Welling, ``Auto-encoding variational Bayes,'' 2013, arXiv:1312.6114. [Online]. Available: https://arxiv.org/abs/1312.6114
\bibitem{b9} S. Liu, ``Wi-Fi Energy Detection Testbed (12MTC),'' 2023, gitHub repository. [Online]. Available: https://github.com/liustone99/Wi-Fi-Energy-Detection-Testbed-12MTC
\bibitem{b10} ``Treatment episode data set: discharges (TEDS-D): concatenated, 2006 to 2009.'' U.S. Department of Health and Human Services, Substance Abuse and Mental Health Services Administration, Office of Applied Studies, August, 2013, DOI:10.3886/ICPSR30122.v2
\bibitem{b11} K. Eves and J. Valasek, ``Adaptive control for singularly perturbed systems examples,'' Code Ocean, Aug. 2023. [Online]. Available: https://codeocean.com/capsule/4989235/tree
\end{thebibliography}

\end{document}